\documentclass[letterpaper, 10 pt, journal, twoside]{IEEEtran}  

\IEEEoverridecommandlockouts                              

\usepackage{amsmath} 

\usepackage{booktabs}
\let\labelindent\relax
\usepackage{enumitem}

\usepackage{cite}
\usepackage{xcolor}
\definecolor{dark-green}{RGB}{12,80,12}

\newcommand{\secref}[1]{Section~\ref{#1}}
\renewcommand{\eqref}[1]{Equation~(\ref{#1})}
\newcommand{\figref}[1]{Figure~\ref{#1}}
\newcommand{\tabref}[1]{Table~\ref{#1}}

\usepackage[breaklinks,colorlinks]{hyperref}
\usepackage{siunitx}
\usepackage[ruled,vlined]{algorithm2e}
\usepackage{algorithmic}
\usepackage{microtype}
\usepackage{flushend}
\SetAlFnt{\small}
\SetAlCapFnt{\small}
\SetAlCapNameFnt{\small}

\newcommand{\website}{\url{http://himos.cs.uni-freiburg.de}}

\usepackage{graphicx}
\usepackage{amsfonts} 

\usepackage{cuted}

\newcommand{\myworries}[1]{\textcolor{black}{#1}}
\newcommand{\myworriestwo}[1]{\textcolor{black}{#1}}

\usepackage[resetlabels]{multibib}
\newcites{S}{References}

\usepackage{tabularx}
\newcolumntype{Y}{>{\centering\arraybackslash}X}
\newcolumntype{Z}{>{\raggedleft\arraybackslash}X}
\usepackage{multirow}

\newcommand{\ours}{HIMOS}
\newcommand{\ourslong}{Hierarchical Interactive Multi-Object Search}
\newcommand{\para}[1]{{\parskip=5pt\noindent\textit{#1}}}

\title{\LARGE \bf
Learning Hierarchical Interactive Multi-Object Search\\for Mobile Manipulation
}

\author{Fabian Schmalstieg$^*$, Daniel Honerkamp$^*$, Tim Welschehold, and Abhinav Valada
\thanks{$^*$These authors contributed equally.}%
\thanks{Department of Computer Science, University of Freiburg, Germany.}%
\thanks{This work was funded by the European Union’s Horizon 2020 research and innovation program under grant agreement No 871449-OpenDR. Toyota Motor Europe supported this project with an HSR robot for experiments.}
}

\begin{document}

\maketitle
\thispagestyle{empty}
\pagestyle{empty}

\begin{abstract}
Existing object-search approaches enable robots to search through free pathways, however, robots operating in unstructured human-centered environments frequently also have to manipulate the environment to their needs. In this work, we introduce a novel interactive multi-object search task in which a robot has to open doors to navigate rooms and search inside cabinets and drawers to find target objects. These new challenges require combining manipulation and navigation skills in unexplored environments. We present \ours{}, a hierarchical reinforcement learning approach that learns to compose exploration, navigation, and manipulation skills. To achieve this, we design an abstract high-level action space around a semantic map memory and leverage the explored environment as instance navigation points. 
We perform extensive experiments in simulation and the real world that demonstrate that\myworries{, with accurate perception, the decision making of }\ours{} effectively transfers to new environments in a zero-shot manner. It shows robustness to unseen subpolicies, failures in their execution, and different robot kinematics. These capabilities open the door to a wide range of downstream tasks across embodied AI and real-world use cases.
\end{abstract}

\section{Introduction}\label{sec:intro}
Autonomous navigation and exploration in unstructured indoor environments require a large variety of skills and capabilities. Pathways may be blocked and objects of interest may be stored away. Thus far, existing multi-object search tasks and methods have focused on environments that can be freely navigated with openly visible target objects~\cite{wani2020multion, fang2019scene, schmalst22exploration}. \myworries{However, as we move to human-centered environments such as households, neither of these assumptions hold.} We introduce a novel \textit{Interactive Multi-Object Search} task in which target objects may be located inside articulated objects such as drawers and closed doors have to be opened to explore the environment. As a result, only navigation is insufficient to accomplish the task and the robotic agent has to physically interact with the environment to manipulate it to its needs. 

Multi-object search tasks pose long-horizon problems with non-trivial optimal policies. Methods such as frontier exploration~\cite{yamauchi1997frontier} offer guarantees to explore the entire environment if given enough time. However, they often do not take the context of the environment into account and result in long far from optimal paths while moving from one frontier point to the next. On the other end of the spectrum, learning-based methods can take unstructured observations into account and have been shown to learn good exploration strategies~\cite{wani2020multion, schmalst22exploration}, but they struggle with the long-horizon nature of the task. Moreover, since the robot also has to interact with the environment, both the action space and task horizon increase even further, and existing exploration methods are insufficient to accomplish the task.\looseness=-1

\begin{figure}[t]
	\centering
  		\includegraphics[width=0.8\linewidth,trim={0.0cm 0.0cm 0.0cm 2.9cm},clip,angle =0]{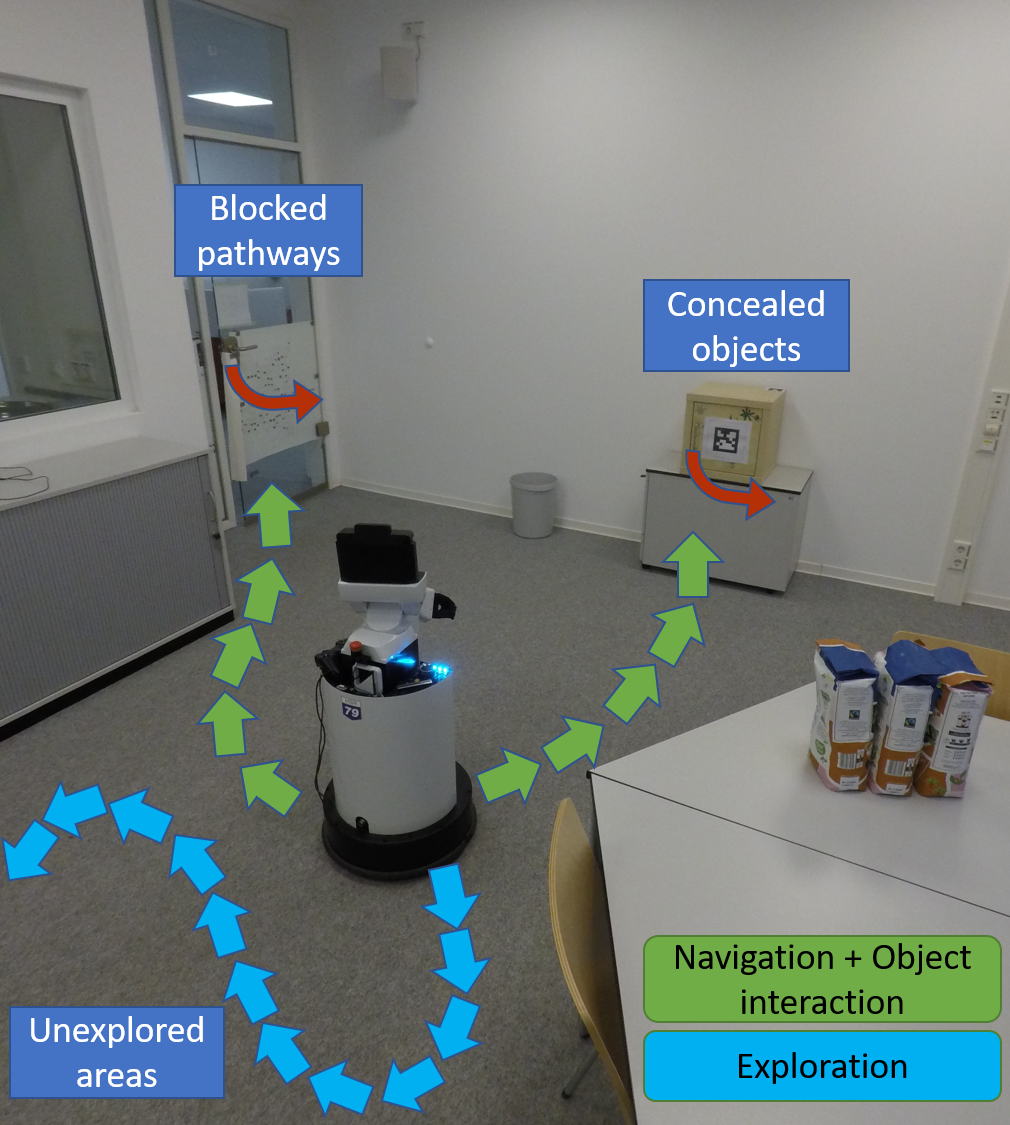}
	\caption{We introduce the  \textit{Interactive Multi-Object Search} task in which an agent has to autonomously search and manipulate the environment to find a set of target objects. To succeed, the agent has to free pathways by opening doors and searching inside articulated objects such as cabinets and drawers.}
  	\label{fig:teaser}
  \vspace{-0.2cm}
\end{figure}

In this work, we propose \ourslong{} (\ours{}), a hierarchical, reinforcement learning\myworries{-based} approach to learn both exploration and manipulation skills and to reason at a high level about the required steps. \myworries{Reinforcement learning has shown to work well in unexplored environments with high-dimensional observation spaces~\cite{chaplot2020learning, chen2020soundspaces}. }We combine learned motions for local exploration in continuous action spaces~\cite{schmalst22exploration} and frontier exploration for long-horizon exploration~\cite{yamauchi1997frontier} together with mobile manipulation skills for object interactions~\cite{honerkamp2022learning}. We use semantic maps as the central memory component, which have shown to be an expressive and sample-efficient representation for these tasks~\cite{schmalst22exploration} and design a high-level action space that exploits the acquired knowledge about the environment. By leveraging explored object instance locations as navigation waypoints, our approach efficiently learns these complex tasks from little data and consistently achieves success rates above 90\% even as the number of target objects increases. By equipping all the low-level skills with mobility, we remove the \myworries{``hand-off''} problem in which subpolicies have to terminate in the initial set of the following skill~\cite{gu2022multi, szot2021habitat}. Lastly, we transfer the trained agent to the real world and demonstrate that it successfully accomplishes these tasks in a real office environment. In particular, we replace the subpolicies from simulation with unseen real-world variations and find that the policy is able to generalize to these unseen subpolicies and is robust to failures in their execution, making it highly modular and flexible for transfer. Finally, we present ablation studies to evaluate the impact of the main design decisions.

To summarize, the following are the main contributions:
\begin{enumerate}
    \item We propose an interactive multi-object search task that requires physical interactions with articulated objects, opening doors, and searching in cabinets and drawers.
    \item We present a hierarchical, reinforcement learning\myworries{-based} approach that combines exploration and manipulation skills based on semantic knowledge and instance navigation points to efficiently solve these long-horizon tasks.
    \item We demonstrate the capabilities of this approach in both simulated and real-world experiments and show that\myworries{, given accurate perception, its decision making} achieves zero-shot transfer to the real world, unseen environments, unseen subpolicies, and is robust to unseen failures.
    \item We make the code for both the task and models publicly available at \website{}. 
\end{enumerate}

\section{Related Work}\label{sec:related}

\para{Object search tasks:} Exploration and the ability to find items of interest is a key requirement for a wide range of downstream tasks. Previous work\myworries{s have} proposed methods to maximize coverage of explored space~\cite{chen2018learning} and to find specific objects of interest based on vision~\cite{chaplot2020learning}, auditory signals~\cite{chen2020soundspaces, younes2021catch} or target object categories~\cite{qiu2020learning, druon2020visual}. In ordered multi-object search tasks, the agent has to find $k$ items in a fixed order in game environments~\cite{beeching2021deep} or realistic 3D apartments~\cite{wani2020multion}. In unordered multi-object search, the agent simply has to find the target objects as fast as possible, irrespective of the order~\cite{fang2019scene, schmalst22exploration}. We focus on this unordered task. As we aim to demonstrate our system on a real robot, we follow Schmalstieg~et~al.~\cite{schmalst22exploration} and use the full continuous action space. This is in contrast to most previous work\myworries{s} which focus on a simplified granular discrete action space.  Existing search tasks assume that the desired goals can be freely reached by the agent. The interactive navigation task~\cite{fei2020interactive} relaxes this assumption by placing objects that the robot has to push away to reach the target. In contrast, we introduce an interactive search task for mobile robots equipped with a manipulator, that requires interaction with articulated objects to clear the path or reveal concealed objects. This requires integrating navigation and manipulation. Lastly, in contrast to most previous work\myworries{s}, we demonstrate that our approach successfully transfers to the real world.

\para{Exploration} requires both understanding and memorizing the seen environment and decision-making to explore the remaining space. Previous work has introduced both implicit and explicit memory mechanisms. Implicit memory agents either learn a direct end-to-end mapping from RGB-D images to actions or store embeddings of previous observations and retrieve them with an attention mechanism~\cite{fang2019scene}. Other methods build explicit maps of the environment by projecting the RGB-D inputs into a global map. Commonly, this map is also annotated with semantic labels~\cite{beeching2021deep, schmalst22exploration, wani2020multion}.  
Further, combining short- and long-term exploration by learning an auxiliary prediction of the direction to the next closest object has proven to result in a strong performance in continuous action spaces~\cite{schmalst22exploration}. 
We use this approach for low-level exploration.
Frontier exploration~\cite{yamauchi1997frontier} samples points on the frontier of the explored space and then navigates to these points. Instead of sampling, Ramakrishnan~et~al.~\cite{ramakrishnan2022poni} predict a potential function towards a target object. SGoLAM~\cite{kim2021sgolam} combines mapping and a goal detection module. If no goal object is detected, it explores with frontier exploration and navigates directly to the goal otherwise. This results in strong results without any learning component. We include a similar exploration strategy as an option in our hierarchical approach. 
\myworriestwo{Zheng et al.~\cite{zheng2022towards, zheng2023system} leverage POMDP-solvers over an explicit belief representation to plan next viewpoints.}

\para{Articulated object manipulation} such as opening doors and drawers requires control of both the base and arm of the robot~\cite{rofer2022kineverse}. Existing approaches often separate both aspects and execute sequential navigation and manipulation. In our evaluation in simulation, we follow this approach and use BiRRT~\cite{qureshi2015intelligent} to generate a motion plan for the robot arm.
Recent work\myworries{s} train an agent to control the base of the robot via reinforcement learning to follow given end-effector motions~\cite{honerkamp2021learning, honerkamp2022learning}. We use this method in our real world evaluation as it generalizes across different robots, tasks, and environments.

\para{Hierarchical methods} introduce layers of abstraction by decomposing the decision-making into higher and lower-level policies. This shortens the time horizon of the Markov decision process (MDP) for the higher-level policies and enables the agent to combine different modules or skills at lower levels. At the same time, joint training of low- and high-level policies is often unstable and hard to optimize~\cite{nachum2018data}. We focus on combining pretrained subpolicies. Joint finetuning of these policies offers a path to further performance improvements in the future. Pretraining is a common approach to increase the stability of the policy in hierarchical reinforcement learning~\cite{hutsebaut2022hierarchical}. 
While naive skill-chaining of arbitrary skills often results in \myworries{``hand-off''} failures in which the subsequent skill cannot start from the current state~\cite{szot2021habitat}, we resolve this issue by adding mobility to all the low-level skills in the real world execution, without the need for region-rewards~\cite{gu2022multi}. A common navigation abstraction is to learn to set waypoints~\cite{krantz2021waypoint, chen2020learning}, however, these often end up as very near points to the agent. In contrast, we propose instance navigation that provides a prior on important locations and action granularity. ASC~\cite{yokoyama2023adaptive} learns to combine navigation and pick skills for given receptacle locations. In contrast, we learn to search in unexplored environments with objects hidden in articulated objects. Alternatively, behavior trees are often used to decompose tasks hierarchically into a tree structure. These trees can either be fully constructed manually or be used in conjunction with a planner~\cite{iovino2022survey}. \myworriestwo{\cite{golluccio2021robotic} learn values of tree nodes for cluttered object relocation. }In contrast to these approaches, our proposed \ours{} learns high-level decision-making with a two-layer hierarchy.

\section{Interactive Multi-Object Search}

We propose an interactive object search task in which a robotic agent with a mobile base and a manipulator arm is randomly spawned in an unexplored indoor environment. The agent receives a goal vector that indicates $k$ objects out of $c$ categories that it has to find. The episode is considered successful if the agent finds all $k$ objects, where an object is considered found when the agent has seen the object and navigated within a distance of $\SI{1.3}{\meter}$ of it. The episode is terminated early if the agent exceeds 1,000 timesteps.
To succeed, the agent has to explore the space while opening doors that block the way and opening the cabinets that contain the target objects. 
This results in very long-horizon tasks with complex shortest paths, as in contrast to previous multi-object search tasks~\cite{fang2019scene, beeching2021deep, wani2020multion, schmalst22exploration}, the agent needs to manipulate its environment to achieve its goals. 

The agent is acting in a goal-conditional Partially Observable Markov Decision Process (POMDP) $\mathcal{M} = (\mathcal{S}, \mathcal{A}, \mathcal{O}, T(s' | s, a), P(o | s), R(s, a, g))$, where $\mathcal{S}$, $\mathcal{A}$ and $\mathcal{O}$ are the state, action, and observation spaces, $T(s' |s, a)$ and $P(o | s)$ describe the transition and observation probabilities and $R(s, a, g)$ is the reward function. The objective is to learn a policy $\pi(a | o, g)$ that maximises the discounted, expected return $\mathbb{E}_\pi[\sum_{t=1}^{T} \gamma^t R(s_t, a_t, g)]$, where $\gamma$ is the discount factor. In each step, the agent receives a visual observation $o$ from an RGB-D and semantic camera, together with its current pose in the environment and a binary vector $g$ indicating the objects it has to find. The true state $s$ of the environment is unknown and can only be inferred from its observations.


\section{\ourslong{}}\label{sec:approach}
\begin{figure}[t]
	\centering
  	\includegraphics[width=1.\linewidth,trim={0.0cm 0.0cm 0.0cm 0.0cm},clip,angle =0]{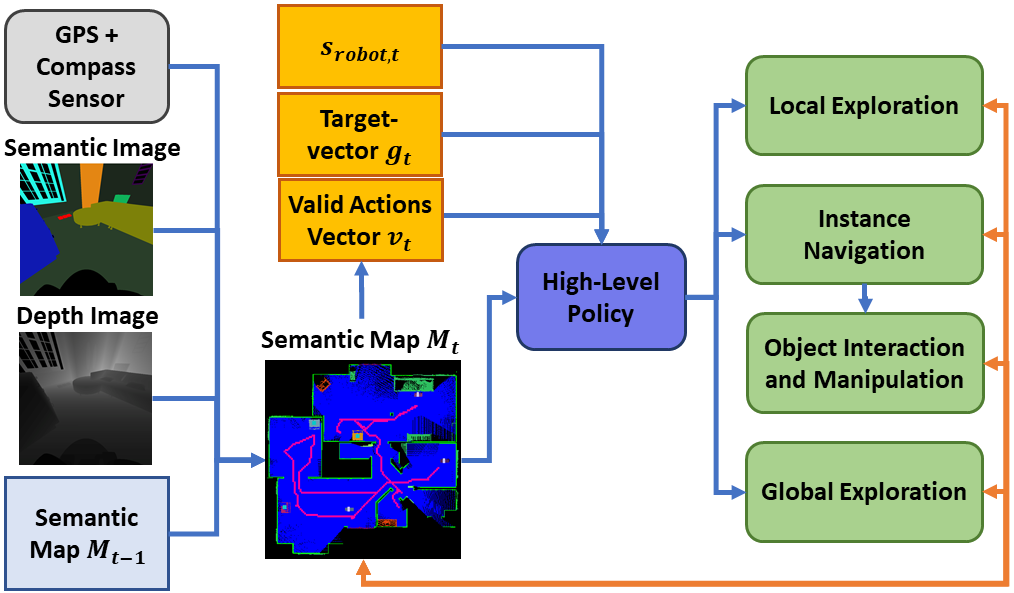}
	\caption{Schematic overview of \ours{}. A semantic map $M_t$ serves as a central memory component and is used and updated across low- as well as high-level modules. This map is extended to a partial panoptic map with instance IDs of relevant objects. Given the remaining objects $g_t$ to find, the robot state $s_{robot, t}$, and the derived valid actions $v_t$, the high-level policy acts in an abstract action space consisting of local and global exploration, navigation to mapped object instances, and a mobile manipulation policy.}
  	\label{fig:architecture}
  \vspace{-0.2cm}
\end{figure}

The challenges introduced by this task require the robot to master an increasing number of different behaviors and skills. To stem this increase in complexity, we propose a hierarchical, reinforcement\myworries{-based} learning approach and design efficient abstractions over states and actions. The model consists of three main components: a mapping module, a set of subpolicies for exploration, navigation, and interaction, and a high-level policy module to select the next low-level action to execute. An overview of the approach is depicted in \figref{fig:architecture}.


\para{Assumptions:} The focus of this work is on learning long-horizon decision-making, exploration, and search. We abstract from real-world perception and assume to have access to the following:
(i)~a semantic camera that produces accurate semantic object labels, (ii)~accurate depth and localization, (iii)~an object detection module to perceive object poses in the environment and to detect whether an object interaction was successful. In training and simulation, the object detection module uses ground truth poses from the simulator while in real-world experiments, we rely on AR markers placed on the respective objects. Interaction failure is detected based on the change in pose of these markers after the interaction.
\myworries{(iv) the environment remains static except for the agent's own interactions with it. For example, objects are not being displaced by other people during the search process.}

\subsection{Mapping Module}
\label{sec:mapping}

A semantic map of the environment serves as a central memory component across all policies and modules. To construct this map, 
we build upon our previously introduced mapping module~\cite{schmalst22exploration}. \myworries{We extend this map to a partial panoptic segmentation~\cite{mohan2022amodal} map by labeling task-specific objects such as doors and cabinets with an instance-specific color, which is randomly assigned whenever a new instance is detected. I.e. an instance's color changes from episode to episode, but remains consistent within an episode. The robot receives the semantic masks and depth at a resolution of $128\times128$ pixels, generates a point cloud from this information, and projects the points into a local top-down map, using the top-most point for each cell. With the local map and its current localization, the agent then updates an internal global map which is further annotated with the agent's trace and encoded into an RGB image. Target objects are mapped as a special color, making the approach agnostic to the underlying classes of the target objects. After the agent has segmented and approached a target object, it updates the object's annotation with a fixed color coding to mark the corresponding object as “found”.} In the simulation, the robot receives the semantic labels from the simulator's semantic camera. In the real world, we simulate access to semantic labels as follows: we pre-build a map of the environment using Hector SLAM~\cite{kohlbrecher2011flexible} and annotate it with labels for target objects, cabinets, and doors. All the other occupied space is mapped to the wall category. At test time, we then use the robot's depth camera to build the same local map as in the simulation and overlay it with this pre-annotated semantic map. 
Localization in the real world is done on the same pre-recorded map. 

In each step, the agent then extracts an egocentric map from the global map and passes two representations of this map to the encoder: a coarse map of dimension $224 \times 224 \times 3$ at a resolution of $\SI{6.6}{\centi\meter}$ and a fine-grained map of dimension $84 \times 84 \times 3$ at a resolution of $\SI{3.3}{\centi\meter}$. I.e., they cover $\SI{14.8}{\meter} \times \SI{14.8}{\meter}$ and $\SI{2.77}{\meter} \times \SI{2.77}{\meter}$, respectively. 

\subsection{Subpolicy Behaviors}
\label{sec:low-level}

In this section, we describe the different subpolicy behaviors that are available to the high-level policy. 

\para{Local Exploration:} Object search requires smart movement for local navigation and efficient exploration around corners and corridors. For this, we use our previously introduced exploration policy~\cite{schmalst22exploration}. The policy receives the semantic map, robot state, and target objects and predicts the direction to the next closest target object. It then communicates this prediction to a reinforcement learning agent which produces target velocities for the base of the robot.
The local exploration policy is pretrained on the multi-object-search task~\cite{schmalst22exploration} in the same train/test split. However, we change the robot to Fetch, adapt the collision penalty from $-0.1$ to $-0.15$, and include open doors in the scenes. After the exploration policy is trained, it is kept frozen during high-level policy training. We adjust the panoptic labels provided by the mapping module on the fly to match the simpler, instance-unaware, semantic map that this subpolicy was trained with. In particular, doors and cabinets are colored as obstacles for the subpolicy. This shields the subpolicy from information that is not required to solve the downstream task. 
During training, when selected by the high-level policy, the exploration policy is executed for four time steps, giving the high-level policy control to quickly react to new knowledge of the environment. During the evaluation, we found it beneficial to execute it for a longer period of 20 time steps.\looseness=-1


\para{Global Exploration:} While the local exploration policy has been shown to produce efficient search behavior, it can struggle to navigate to faraway areas. This, however, is a strength of frontier exploration~\cite{yamauchi1997frontier}, which samples points at the frontier to (often far away) unexplored areas. While frontier exploration on its own can lead to long inefficient paths, the high-level policy can learn to select the appropriate exploration strategy for the current context. In each iteration, the sampled frontier point is drawn onto the map, allowing the high-level policy to observe where it would navigate to before deciding which subpolicy to execute. Frontier points that lie outside the range of the agent's egocentric map are projected onto a circle around the agent and marked in a different color, indicating that it is potentially a long-distance navigation. If selected by the high-level policy, the agent uses its navigation policy described below to navigate to the frontier point.

\para{Instance Navigation:} Learning navigation at the right level of abstraction can be challenging. Approaches such as setting waypoints are often difficult to optimize or decay to only selecting nearby points, removing the benefits of the abstraction. Instead, we leverage the acquired knowledge about the environment by using object instances as navigation points: the high-level policy can directly navigate to the discovered object instances by selecting their instance ID (for simplicity restricted to target objects, doors, and cabinets). We implement this action space as a one-hot encoding that maps to instance colors on its map (this assumes a maximum number of instances). We furthermore find it beneficial for learning speed~\cite{huang2020closer} to only allow the agent to navigate to doors or cabinets that have not been successfully opened yet (cf. invalid action masking below).
While less fine-grained than arbitrary waypoints, this results in an efficient set of navigable points across the map that, as we demonstrate in \secref{sec:experiments}, is well optimizable and results in a strong final policy.\looseness=-1

The respective navigation goals 
are set to a pose slightly in front, or for the goal objects directly to the detected pose of the corresponding object. This navigation goal is then fed to an $A^*$-planner which produces a trajectory at a resolution of \SI{0.5}{\meter}. For training speed, we do not execute the full path in simulation, but set the robot's base pose to the generated waypoints and only collect observations from these points. In the real world, we use the ROS navigation stack to move the robot to the goal. The policy may fail in some situations, for example, due to collision with obstacles or narrow doorways. In this case, the agent returns to the last feasible waypoint, and given the updated semantic map $M_t$, the high-level policy has to make a new decision.

\para{Object Interaction and Manipulation:} If the high-level policy chooses to navigate to a closed door or cabinet, this automatically triggers an interaction action that is executed once the navigation has been successfully completed. For fast simulation, we train the agent with magic actions that either open the object successfully or fail with a probability of 15\% and leave the object untouched, in which case the agent has to decide whether to try to open it again. The training with failure cases enables it to learn a re-trial behavior to recover from failed attempts.

At test time, the agent has to physically execute the interactions. In the simulation, we replace the interaction subpolicy with a BiRRT motion planner~\cite{qureshi2015intelligent} and inverse kinematics to execute a push-pull motion. The success of these motions depends on the pose of the robot and the object. Implementation details can be found in the supplementary material. In the real world, we replace these subpolicies with the $N^2M^2$ mobile manipulation policy~\cite{honerkamp2022learning}. Given the pose of the object handle (based on an AR marker) and the object label (door, drawer, or cabinet), it generates end-effector motions learned from demonstrations to open the object together with base commands that ensure that these motions remain kinematically feasible. The model is pretrained without additional retraining or finetuning. The agent returns a success indicator to the high-level policy and in case of failure, the agent again has to decide whether to repeat the interaction. 

\subsection{High-level Decision Making}\label{sec:high-level}

Efficient high-level decision making requires the right level of abstractions of states and actions. We hypothesize that object- and instance-level decision making is such an efficient level of abstraction for embodied search tasks. We design a high-level policy around this idea. In particular: 
(i)~We propose an instance navigation subpolicy that leverages the agent's accumulated knowledge about the environment. It provides a prior on important places and on the granularity of navigation points, making it data-efficient and well-optimizable. 
(ii)~As objects are discrete instances, the resulting full action space remains discrete, avoiding the complexities of mixed action spaces. At the same time, all the subpolicies still act directly in continuous action spaces, allowing for direct transfer to real robotic systems.
(iii)~We abstract from reasoning about exact robot placements in the real world by shifting the responsibility of mobility into the subpolicies. This ensures that the subpolicies can start from a large set of initial positions, resolving the \myworries{``hand-off''} problem from naive skill-chaining~\cite{szot2021habitat} and strongly simplifies the learning process for the high-level policy. Furthermore, it enables us to change out the subpolicies to unseen subpolicies in the real world.
(iv)~We incorporate subpolicy failures into the training process, enabling the high-level policy to learn a retrial behavior if execution fails.

\para{\myworries{Action Space:}} The high-level policy acts in a Semi-Markov Decision Process (SMDP) in which the actions model temporarily extended behaviors and act at irregular intervals~\cite{sutton1991between}. \myworries{It makes a decision whenever the last invoked subpolicy returns, successful or not.}
The high-level action space consists of (i)~invoking the local exploration policy~\cite{schmalst22exploration}, (ii)~invoking global frontier exploration~\cite{yamauchi1997frontier}, (iii)~instance navigation with subsequent object interaction (if available), 
where instance IDs are selected through a one-hot vector mapping to fixed colors. The current task instantiation assumes a maximum of ten instances per episode, resulting in an overall 12-dimensional discrete action space. 


\para{Adaptive Discounting:} 
The high-level policy acts at irregular intervals, as the duration of the subpolicies varies largely. We correct for this time bias and accurately reflect the long-term consequences of actions with adaptive discounting. 

\para{Invalid Action Masking:} The availability of high-level actions varies with the state of the environment, e.g. navigating to an object instance depends on the instance being mapped, opening a cabinet is only possible if it is mapped and closed. 
We infer a valid actions vector $v_t$ from the agent's observations and
incorporate it into the training by masking out invalid actions as well as including it in the observation space of the agent. 
As a result, the agent can learn more effectively, speeding up the training process~\cite{huang2020closer}. We implement this by replacing the logits of the invalid actions with a large negative number.


\para{\myworries{Observation Space and }Architecture:} 
\myworries{The observation space of the high-level policy consists of the coarse semantic map $\in \mathcal{R}^{224 \times 224 \times 3}$, the $c$-dimensional target object vector $g_t$ and the $i$-dimensional binary valid actions vector $v_t$ (see invalid action masking), where $i$ is the number of instances, and a 22-dimensional robot state vector $s_{robot, t}$. The robot state consists} of linear and angular base velocities, the last low-level actions, sum of collisions over the last ten steps, a current collision flag, and a normalized history over the last 16 high-level actions taken.
\myworries{Following previous work~\cite{chen2018learning, schmalst22exploration}, the agent first encodes the coarse semantic map with a ResNet-18~\cite{he2016deep}. Then it concatenates these features with the other, structured observations.}
The high-level policy is trained with Proximal Policy Optimization (PPO)~\cite{schulman2017proximal}. 
We report hyperparameters\myworries{and full network} architectures in the supplementary material.


\para{Reward Functions:} The high-level policy is trained with the accumulated rewards of the invoked subpolicies. The subpolicies collect the following rewards:
(i)~A sparse positive reward of +10 for finding a target object,
(ii)~A sparse positive reward of +3 for opening a door,
(iii)~A penalty of -0.1 per collision for navigation policies,
(iv)~A negative traveled distance reward to encourage the high-level policy to find efficient compositions of the subpolicies. As the navigation policy does not get physically executed during training, we set it to $-0.05$ for each invocation of the local exploration policy and to $-0.05 * \text{number of waypoints}$ for the navigation policy. This results in a similar penalty per distance traveled. 

\section{Experimental Evaluations}\label{sec:experiments}
We extensively evaluate our approach both in simulation and real-world experiments. We aim to answer the following:\\
I) Does the high-level policy learn to make decisions that lead to efficient exploration of the environment, improving over alternative decision rules?\\
II) What is the impact of the different subpolicies, in particular the local and global exploration policies?\\
III) Does the learned behavior transfer to the real world and to execution with different subpolicies?\\
IV) Is the overall system capable of successfully solving extended tasks involving many physical interactions within a single episode in the real world?

\subsection{Experimental Setup}

We instantiate the task in the iGibson simulator~\cite{chengshu2021igibson2}\myworries{, which builds on the PyBullet physics engine}. Each scene contains three cabinets placed randomly across a set of feasible locations. All the doors in the scene are initially in a closed state. We then construct tasks of finding 1-6 target objects, matching the hardest setting in previous work \cite{wani2020multion, schmalst22exploration}. We randomly place up to three target objects across the free space of the entire apartment and up to three objects inside the cabinets. We use the same eight training scenes as the iGibson challenge and use the remaining seven apartments for evaluation. 
The embodied agent is a Fetch robot, equipped with a mobile base with a differential drive, a height-adjustable torso, and a 7-DoF arm. We scale the robot's size by a factor of 0.85 to navigate the narrow corridors of all apartments. Its raw 10-dimensional action space consists of a continuous linear and angular velocity for the base together with the torso and arm-joint velocities. The robot is equipped with an RGB-D camera with a field of view of 79 degrees and a maximum depth of \SI{5.6}{\meter}.

\subsection{Baselines}

We compare our approach against different high-level decision-making modules and ablations of the action space.\\
\para{Greedy:} A greedy high-level decision-making strategy that immediately drives to any newly mapped task object (door, cabinet, target object) if available and otherwise selects either the local or global exploration policies with equal probability. \\
\para{SGoLAM+:} SGoLAM~\cite{kim2021sgolam} combines non-learning based approaches to achieve very strong performance on the CVPR 2021 MultiOn challenge. It explores the map with frontier exploration until it localizes a target object, then switches to a planner to navigate to the target. We reimplement the author's approach for continuous action spaces and directly use the semantic camera for goal localization which further improves the performance. We then modify the action execution to open doors and cabinets when applicable.\\ 
\para{\ours{}:} The hierarchical approach presented in \secref{sec:approach}.\\ 
\para{w/o frontier} removes global exploration from the subpolicy set.
\para{w/o expl} removes local exploration from the subpolicy set.\\
\para{w/o IAM} removes the invalid action masking and instead penalizes the agent with -2.5 for selecting invalid actions.

\para{Metrics:} We evaluate the models' ability to find all the desired objects using the success rate and we evaluate the optimality of the search path with the success-weighted path length (SPL)~\cite{anderson2018evaluation}. In the simulation, we evaluate 25 episodes per scene, the number of target objects, 
and report the average over three random training seeds.
This results in a total of $ 25\cdot 6\cdot 8\cdot 3 = 3600$ episodes for seen and $ 25 \cdot 6\cdot 7 \cdot 3 = 3150$ episodes for unseen apartments for each approach.

\subsection{Simulation Experiments}
\begin{table*}[t]
    \centering
    \caption{Evaluation of seen and unseen environments, reporting the success rate (top) and SPL (bottom).}
    \label{tab:seen_combined}
    {\color{black}
    \begin{tabularx}{\textwidth}{cl|YYYYYY|Y||YYYYYY|Y}
      \toprule
        & Model & \multicolumn{7}{c||}{Seen} & \multicolumn{7}{c}{Unseen}\\
        \cmidrule{3-16}
        & & 1-obj & 2-obj & 3-obj &4-obj & 5-obj & 6-obj & Avg 1-6 & 1-obj & 2-obj & 3-obj &4-obj & 5-obj & 6-obj & Avg 1-6 \\
      \midrule
        \parbox[t]{3mm}{\multirow{6}{*}{\rotatebox[origin=c]{90}{Success}}} 

        & Greedy &  \textbf{96.8} & 94.0 & 91.0 & 91.3 & 89.7 & 88.0 & 91.8 & 96.4 & 94.9	& 92.6	& 93.7	& 91.4	& 87.2	& 92.7 \\


        & SGoLAM+ &  96.0 & 93.7 & 90.2 & 85.8 & 88.0 & 85.3 & 89.8 & 95.6	&94.1	&94.7	&90.5	&91.6	&91.6	&93.0\\
        

        & w/o frontier &  81.1 & 68.9 & 59.1 & 68.5 & 71.9 & 35.7 & 64.2 & 82.2	&74.2	&65.0	&61.8	&65.3	&50.5	&66.5\\
        
    
        & w/o expl &  93.2 & 88.4 & 87.2 & 91.6 & 87.6 & 58.4 & 84.4 & 93.4	&90.2	&85.8	&81.8	&70.8	&79.6	&83.6\\


        & w/o IAM &  95.1 & 95.1 & 90.3 & 89.5 & 84.0 & 87.2 & 90.2 & 96.5	&94.3	&92.1	&82.7	&85.4	&78.2	&88.2\\


        & \ours{} &  \textbf{96.8}	&\textbf{96.5}	&\textbf{94.0}	&\textbf{94.8}	&\textbf{93.5}	&\textbf{91.7}	&\textbf{94.6} & \textbf{97.5}	&\textbf{97.9}	&\textbf{95.6}	&\textbf{95.6}	&\textbf{93.5}	&\textbf{93.5}	&\textbf{95.6} \\
        
        \midrule
        \parbox[t]{3mm}{\multirow{6}{*}{\rotatebox[origin=c]{90}{SPL}}}

        & Greedy &  43.1 & 45.3 & 48.6 & 50.8 & 54.1 & 56.9 & 49.8 &45.0	&49.5	&51.8	&57.9	&60.6	&62.0	&54.5\\
        

        & SGoLAM+ &  41.5 & 44.3 & 47.8 & 48.9 & 52.5 & 55.3 & 48.4 & 44.8	&49.1	&54.2	&56.2	&60.3	&64.2	&54.8\\
        

        & w/o frontier &  41.0 & 41.2 & 42.9 & 44.9 & 58.0 & 47.4 & 45.9 &32.0	&45.2	&50.9	&46.3	&49.6	&47.2	&45.2\\


        & w/o expl &  43.3 & 45.7 & 45.4 & 41.6 & 48.8 & 44.9 & 44.9 & 40.9	&43.1	&56.1	&54.7	&45.7	&52.3	&48.8\\


        & w/o IAM &  46.9 & 48.5 & 51.9 & 38.7 & \textbf{61.0} & 51.2 & 49.7 & 45.9	&46.3	&45.2	&55.0	&54.1	&52.3	&49.8\\
        

        & \ours{} &  \textbf{49.7}	&\textbf{51.5}	&\textbf{55.2}	&\textbf{57.5}	&59.5	&\textbf{60.9}	&\textbf{55.7} & \textbf{52.2}	&\textbf{56.0}	&\textbf{57.6}	&\textbf{63.5}	&\textbf{65.7}	&\textbf{68.7}	&\textbf{60.6}\\
      \bottomrule

    \end{tabularx}
    }
\end{table*}

\setlength{\tabcolsep}{1pt}
\renewcommand{\arraystretch}{1}
\begin{figure}[t]
	\centering
	{\setlength{\fboxsep}{0pt}%
  \setlength{\fboxrule}{0pt}%
	\resizebox{.8\linewidth}{!}{%
  	\begin{tabular}{cc}
  		\fbox{\includegraphics[height=0.05\textheight,trim={2cm 2.5cm 2cm 1cm},clip,angle =0]{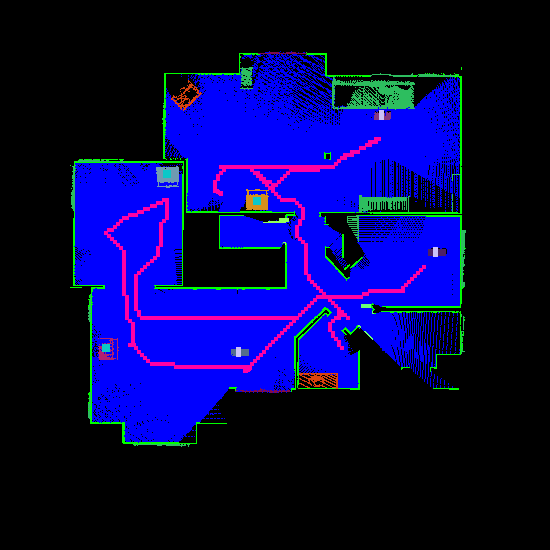}} &
  		\fbox{\includegraphics[width=0.05\textheight,trim={2cm 2.0cm 8.0cm 0cm},clip,angle =90]{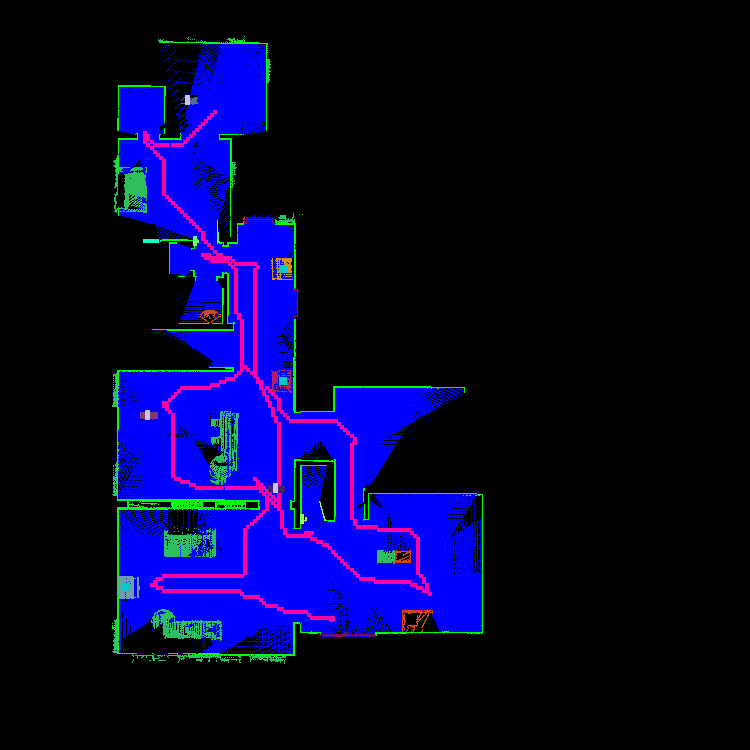}}\\
        \fbox{\includegraphics[height=0.05\textheight,trim={7cm 4.0cm 6cm 4.8cm},clip,angle =0]{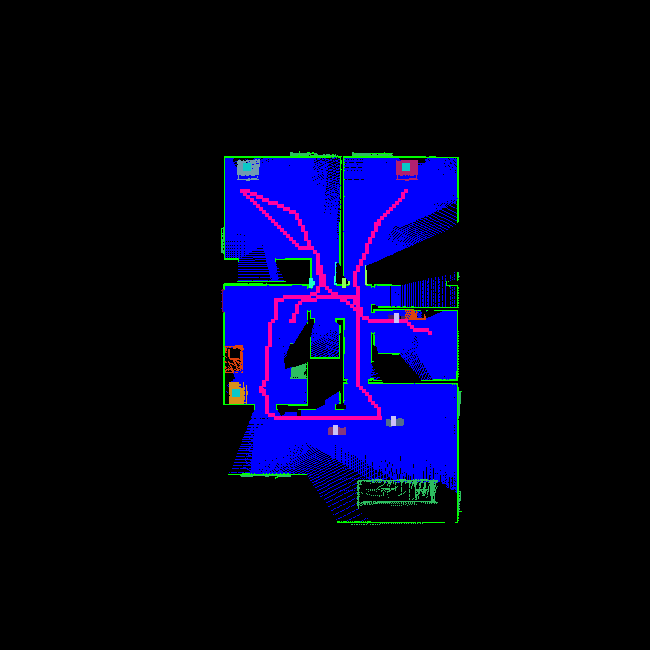}} &  
        \fbox{\includegraphics[width=0.05\textheight,trim={2cm 2.0cm 6.0cm 0cm},clip,angle =90]{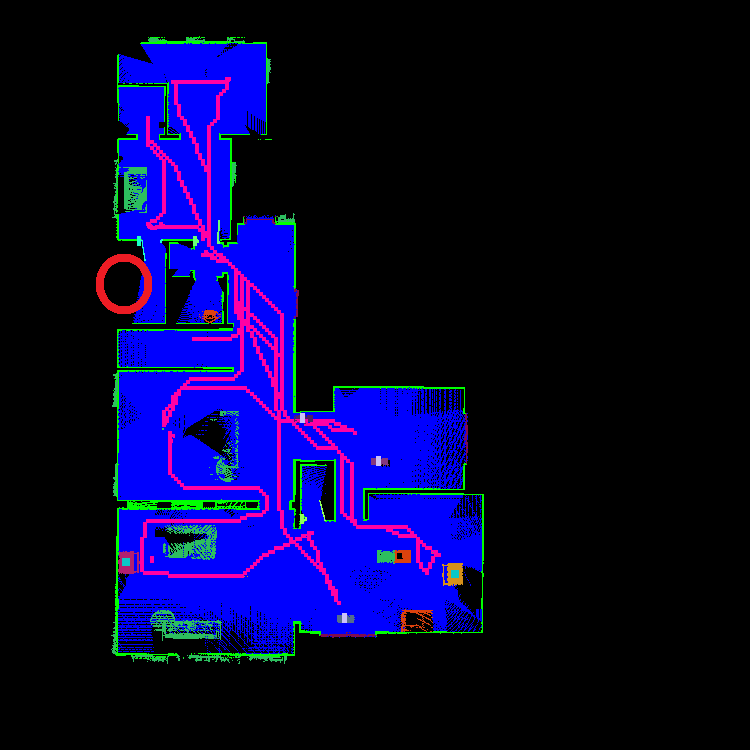}}\\  
	\end{tabular}
	}}
	\caption{Example trajectories of \ours{} in unseen apartments. Black: unexplored, blue: free space, green: walls, red: agent trace, grey: (found) target objects, other colors: miscellaneous objects. Bottom right: the agent failed to find the last object, marked by the red circle, in the given time.}
  	\label{fig:example_trajectories}
     \vspace{-0.2cm}
\end{figure}
\setlength{\tabcolsep}{6pt}
\renewcommand{\arraystretch}{1}

To test the models' abilities to learn to complete the tasks, we first evaluate them in the seen apartments for variable numbers of target objects. The results are reported in \tabref{tab:seen_combined} \myworries{(left)}.
We find that all the compared models achieve good success rates. All three high-level decision-making variations, greedy, SGoLAM+, \ours{}, are able to make reasonable decisions, demonstrating the benefits of the design of the high-level abstractions discussed in \secref{sec:high-level}. Furthermore, our proposed method, \ours{}, further improves over the baselines, consistently achieving the highest success rate and the most efficient paths, as measured by the SPL metric.

We then evaluate the models in the unseen apartments. Note that neither the low- nor high-level policies have seen these scenes during training. The results are shown in \myworries{\tabref{tab:seen_combined} (right)}. We find that the models learned to generalize without any clear generalization gap. The performance is even slightly higher than on the seen apartments, this is in accordance with previous observations~\cite{schmalst22exploration}. This may be due to the validation split containing potentially simpler scenes than the training scenes. 
The evaluations on unseen scenes confirm the observations from the training scenes: \ours{} consistently achieves the highest success rate and the best SPL across all the numbers of target objects. Finally, we find that our hierarchical approach scales very well to longer scenarios, with a very small drop in success rates as the number of target objects increases. To find all six objects, the agent often has to explore the majority of the apartments and interact with a large number of objects.

\subsection{Ablation Study}

\para{Exploration Subpolicies:} To evaluate the impact of the exploration policies, we compare \ours{} to \textit{w/o frontier} and \textit{w/o expl}. We find frontier exploration to have a large impact on success rates. Removing this component reduces the success rate to \myworries{66.5}\%. Removing the local exploration policy leads to a smaller, but nonetheless significant drop of \myworries{12.0} ppt in average success rates as well as a clear drop in SPL. This indicates the different strengths of the two exploration behaviors, as well as that \ours{} learned to use the local exploration policy to increase search efficiency. Again, we find this effect to be consistent across both seen and unseen apartments.

\para{Invalid Action Masking:} We observe that removing the invalid action masking also leads to a drop in both success rates and SPL. Furthermore, we found that masking improves convergence speed by up to 2.5 times. Note that the action masking does not use any further privileged information beyond our perception assumptions (\secref{sec:approach}) of inferring object states.\looseness=-1

Qualitatively, we find that the agent learned sensible behaviors for the task at hand. \figref{fig:example_trajectories} depicts example episodes in the unseen scenes. The agent learned to frequently invoke the local exploration policy while the apartment is still largely unexplored, to then use the global exploration policy to navigate to unexplored corners where target objects could be hidden. Generally, areas further away are being used in order to travel faster to certain areas of interest. The high-level policy also frequently uses frontier-based navigation when the exploration policy is stuck in some area. When a target object lies on the way to another relevant navigation point (a frontier, cabinet, or door), the high-level policy learned to navigate directly to the latter, instead of sequentially navigating to the target object and then proceeding. This saves time and improves efficiency.


\subsection{Real World Experiments}
\begin{table}[t]
    \centering
    \caption{Real world experiments on the HSR Robot.}
    \label{tab:real_world}
    \begin{tabularx}{\columnwidth}{l|YYYYYY|Z}
      \toprule
        Model & 1-obj & 2-obj & 3-obj &4-obj & 5-obj & 6-obj & Total\\
      \midrule
        \textbf{Success}    & 5 & 4 & 4 & 3 & 4 & 3 & \textbf{23}\\
        Collision           & 0 & 0 & 1 & 0 & 0 & 1 & 2 \\
        Interaction failure & 0 & 1 & 0 & 0 & 1 & 1 & 3 \\
        Navigation failure  & 0 & 0 & 0 & 2 & 0 & 0 & 2 \\
        Total Episodes      & 5 & 5 & 5 & 5 & 5 & 5 & \textbf{30}\\
      \bottomrule
    \end{tabularx}
\end{table}

\setlength{\tabcolsep}{1pt}
\renewcommand{\arraystretch}{1}
\begin{figure}[t]
	\centering
	{\setlength{\fboxsep}{0pt}%
  \setlength{\fboxrule}{0pt}%
	\resizebox{.9\linewidth}{!}{%
  	\begin{tabular}{cc} 
    \multicolumn{2}{c}{\includegraphics[width=0.4\linewidth,trim={0cm 0cm 0cm 0cm},clip,angle =270]{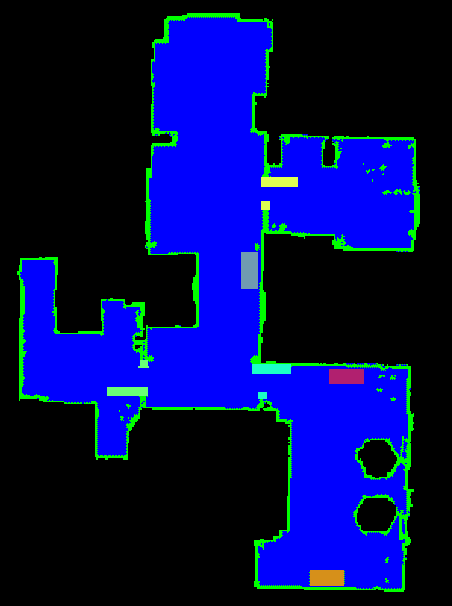}} \\
        \includegraphics[width=0.5\linewidth,trim={13cm 0cm 0cm 0cm},clip,angle =0]{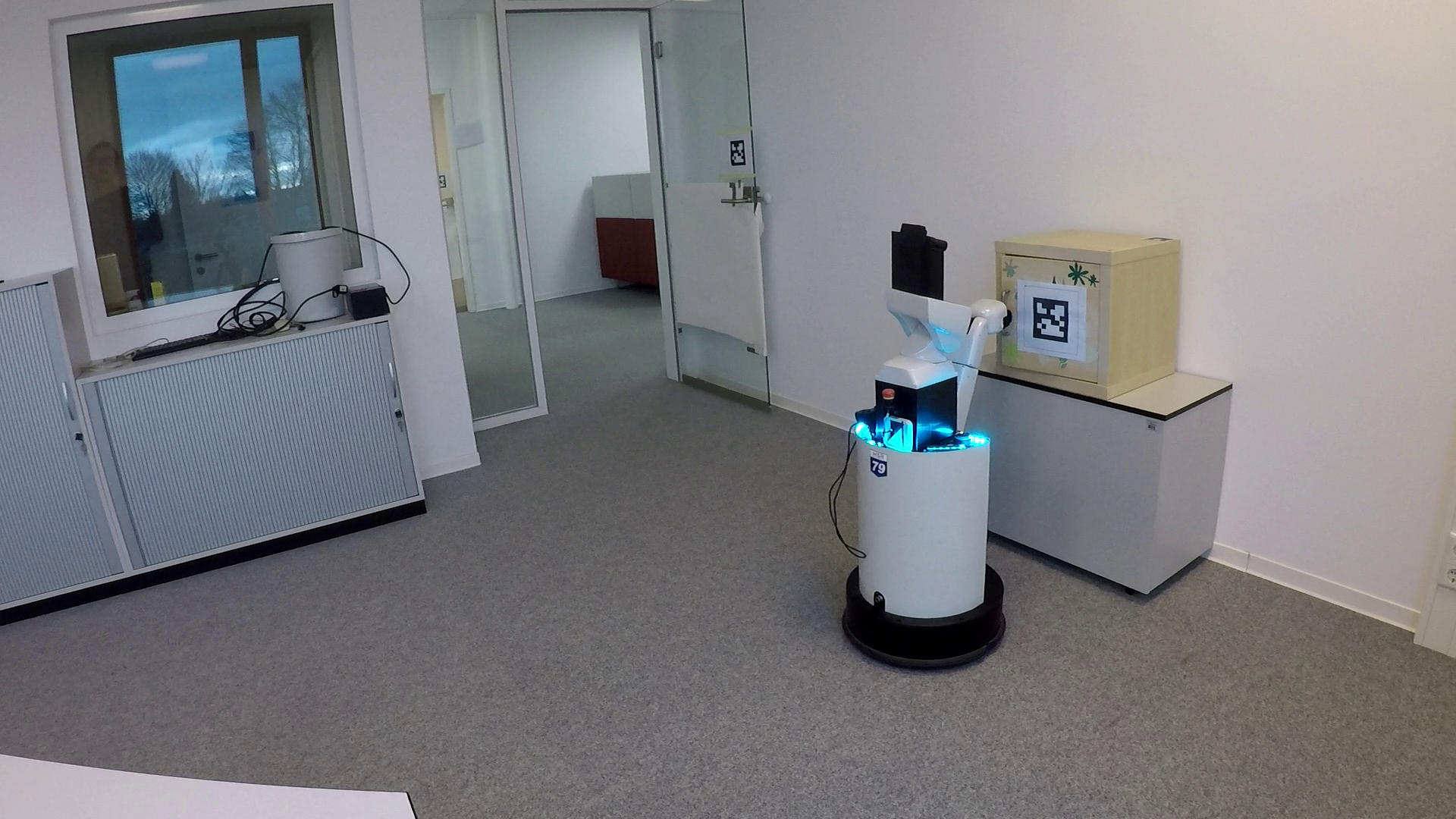} &
  		\includegraphics[width=0.5\linewidth,trim={13cm 0cm 0cm 0cm},clip,angle =0]{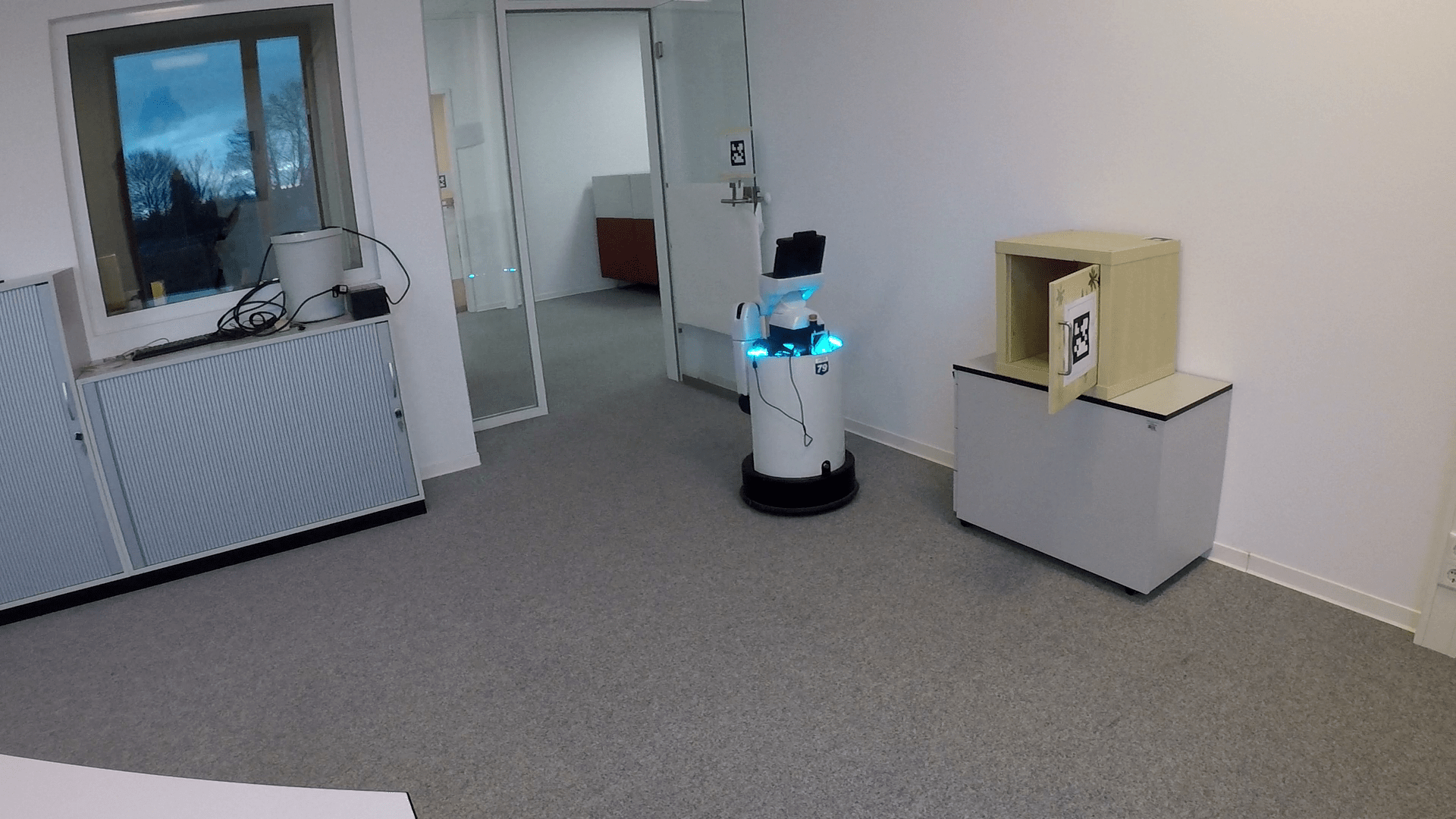} \\
  		\includegraphics[width=0.5\linewidth,trim={6.5cm 0cm 6.5cm 0cm},clip,angle =0]{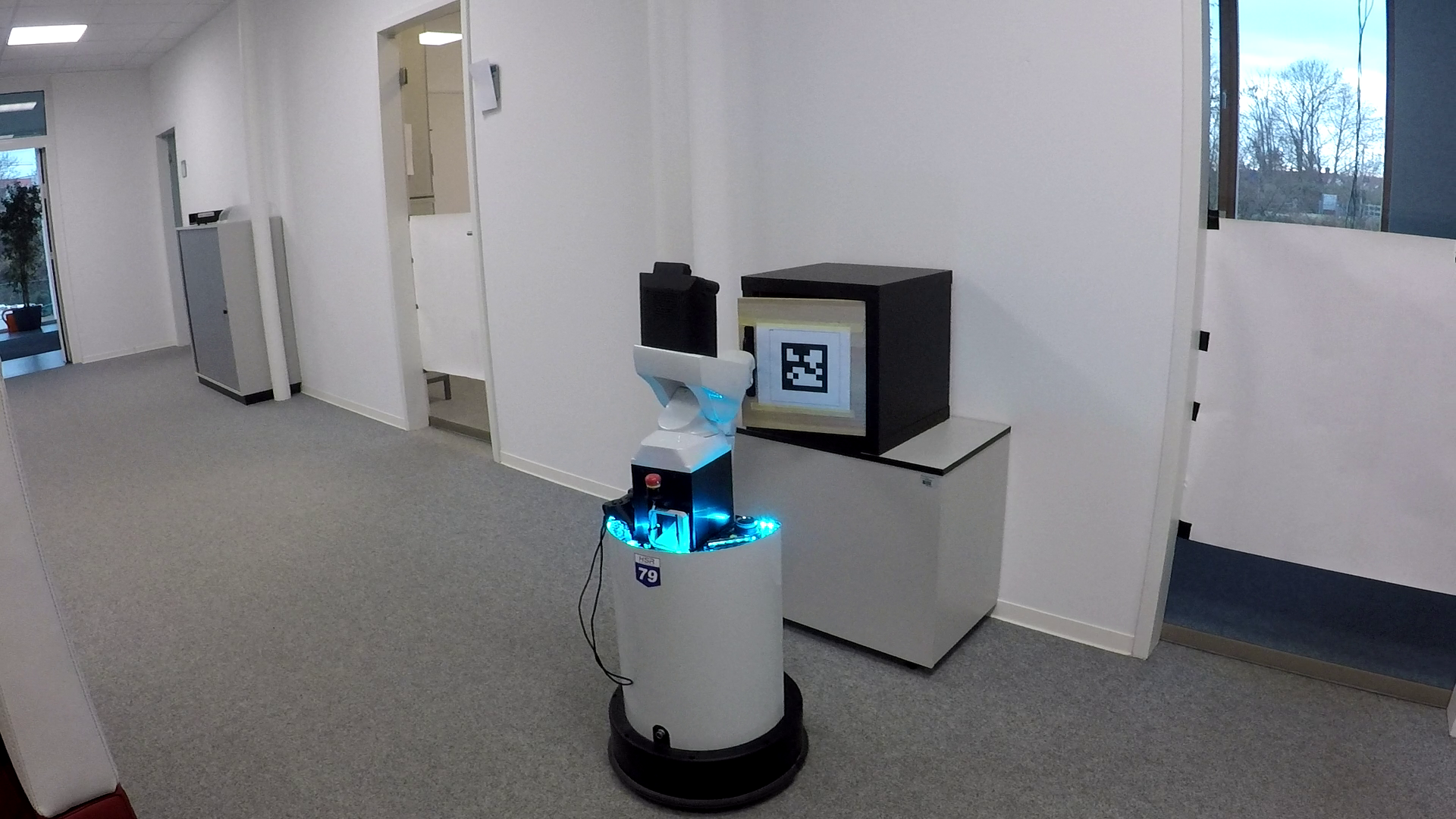} &
        \includegraphics[width=0.5\linewidth,trim={11.5cm 0cm 1.5cm 0cm},clip,angle =0]{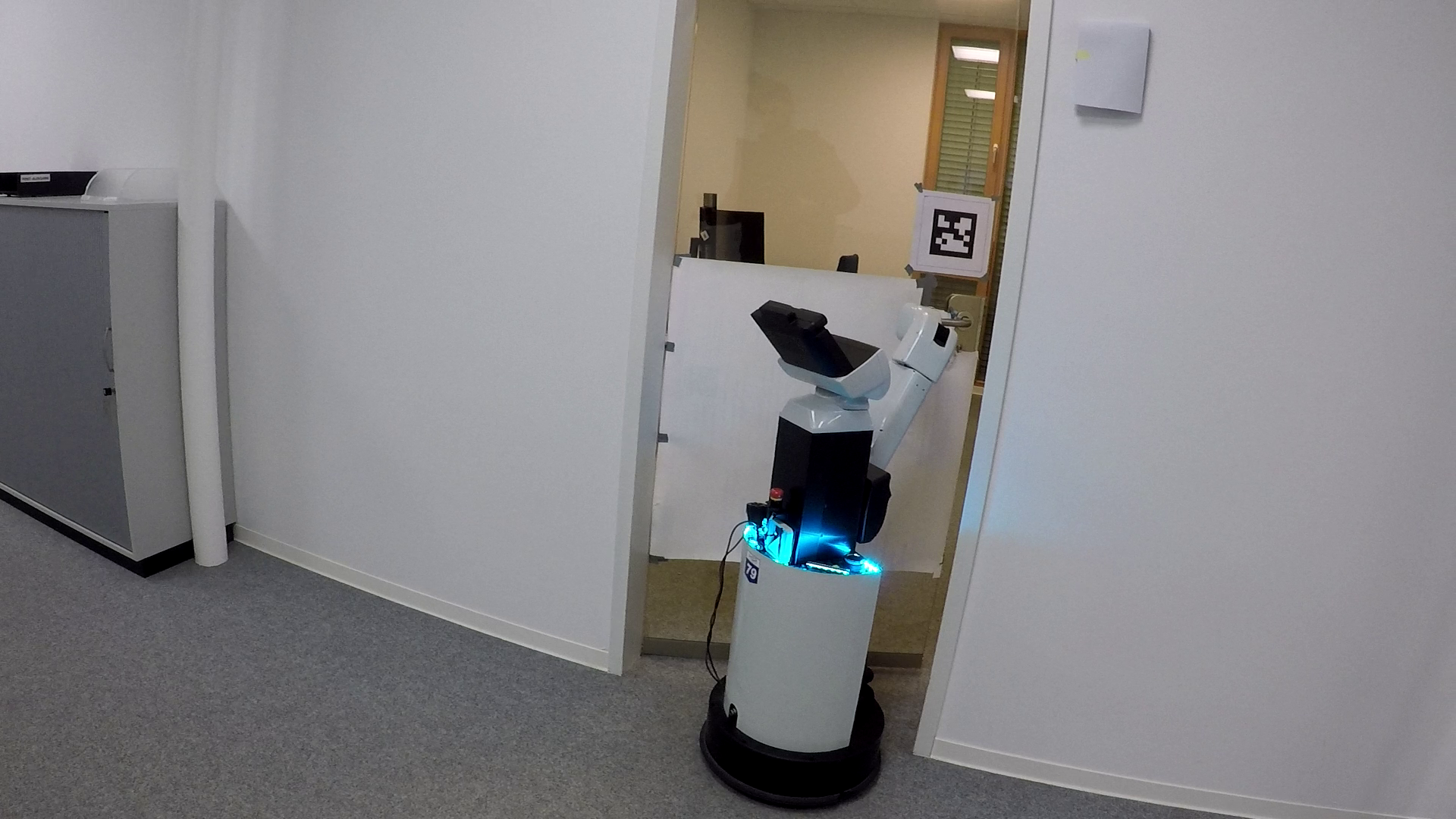}\\  
	\end{tabular}
	}}
	\caption{Top: map of the real world environment. Initial door state and cabinet positions are randomized between episodes. Below: example trajectory in the real world. From top left to bottom right: the agent decides to look inside a target object, then navigates to the hallway, opens a different cabinet and finally opens and drives through a closed door.}
  	\label{fig:real-world}
     \vspace{-0.2cm}
\end{figure}
\setlength{\tabcolsep}{6pt}
\renewcommand{\arraystretch}{1}
We transfer the trained policy to a \textit{Toyota HSR} robot. The robot has a height-adjustable torso and a $5$-\myworries{DoF} arm for environment interactions. It is equipped with an RGB-D camera used for mapping and a $2$D lidar employed for localization in the pre-built map and object avoidance by the ROS navigation stack.
Both the local exploration policy and the high-level behavior policy are transferred to this real-world setting without any further retraining or fine-tuning. The agent requires only minor adjustments to account for the differences in robot geometry and subpolicies. 
See~\secref{sec:low-level} and the supplementary material for details on the real-world subpolicies. 

The experiments are performed in an office building covering three rooms connected by a hallway with a total of three doors. The operation space covers roughly $180$ square meters. We place three articulated objects, two cabinets with a revolute door, and one drawer in the environment. These objects are never seen during training. We evaluate runs with 1-6 target objects in five different scenarios, for a total of 30 episodes. Each scenario defines new positions for the three articulated objects. We randomly chose which doors start in an open or closed state, and start each episode from the room that the last episode terminated in. We cover glass doors to prevent the agent from directly looking through them. 
This evaluation tests (i)~the generalization abilities of both low- and high-level behaviors to the real world\myworries{, given accurate semantic perception}, (ii)~the generalization abilities of the local exploration policy and high-level policy to a different robot, and (iii)~the high-level policy's generalization to unseen subpolicies, as we change both the navigation and manipulation modules. This is an important ability for transfers to different robot models and execution requirements. Lastly, (iv)~the map representation enables easy transfer to different objects as it only requires a mapping to known semantic and instance colors.
\myworries{Our aim is to evaluate the transfer of decision making ability within our stated assumptions. We leave full system evaluation with integrated perception pipeline for future work. We provide details on how this could be implemented in the supplementary.}

The results of the experiments are summarized in Table~\ref{tab:real_world} and example episodes are shown in \figref{fig:real-world} as well as in the supplementary video. The agent successfully completes 76.7\% of the episodes, requiring long sequences of autonomous navigation and physical interactions. The high-level policy proves robust to failures in the subpolicies. These include navigation failures if the planner does not find a valid path to a frontier point and manipulation failures in which the mobile manipulation skill fails to grasp the handle of an articulated object. In this case, the agent is capable of re-triggering the interactions after detecting the failure. A few irrecoverable failures occurred: reaching a safety limit of the wrist joint during door opening, base collisions, and in two cases repeated failures of the navigation stack.


\section{Conclusion}\label{sec:conclusion}
We introduced the interactive multi-object search task in which the agent has to manipulate the environment in order to fully explore it, resembling common household settings. We proposed a novel hierarchical, reinforcement learning\myworries{-based} approach capable of solving this complex task in both simulation and the real world. By combining a high-level policy on abstract action spaces with low-level robot behaviors, we are able to perform long-term reasoning while acting in continuous action spaces. Our approach decouples the perception from decision making which allows a seamless transition to unknown and real-world environments on a differing embodiment. \myworries{Tighter integration with perception and active perception in the lower level are promising areas for future work.} In extensive experiments, we demonstrated the capabilities of our approach and the importance of the individual components in ablation studies.
In future work, we will investigate the benefits of jointly training the high- and low-behavior and integrate more sophisticated mapping modules directly based on the robot sensors.
Further, additional low-level behaviors could extend environment interaction options or perform more goal-oriented active perception actions. \myworries{Detection of more finegrained failure-feedback could further help the high-level decision making. Lastly, the current instance navigation assumes a pre-specified maximum number of possible instances. Regression-based instance selection is a promising avenue to further scale this approach.}



\footnotesize
\bibliographystyle{IEEEtran}
\bibliography{biblio}

\clearpage
\renewcommand{\baselinestretch}{1}
\setlength{\belowcaptionskip}{0pt}

\begin{strip}
\begin{center}
\vspace{-5ex}
\textbf{\LARGE \bf
Learning Hierarchical Interactive Multi-Object Search\\\vspace{0.5ex}for Mobile Manipulation} \\
\vspace{3ex}

\Large{\bf- Supplementary Material -}\\
\vspace{0.4cm}
\normalsize{Fabian Schmalstieg$^*$\hspace{1cm} Daniel Honerkamp$^*$ \hspace{1cm} Tim Welschehold\hspace{1cm} Abhinav Valada}\\
\end{center}
\end{strip}

\setcounter{section}{0}
\setcounter{equation}{0}
\setcounter{figure}{0}
\setcounter{table}{0}
\setcounter{page}{1}
\makeatletter

\let\thefootnote\relax\footnote{$^*$These authors contributed equally.\\Department of Computer Science, University of Freiburg, Germany.\\
Project page: \url{http://himos.cs.uni-freiburg.de}
}%
\normalsize 
In this supplementary material, we provide additional details on the environment setup, the subpolicies as well as details on the training and architecture. Examples of the learned behavior are included in the video material.

\section{Environment Details}
\label{sec:env_details}

\begin{figure}
	\centering
	\resizebox{\columnwidth}{!}{%
  		\includegraphics[width=0.32\columnwidth,trim={0.0cm 0.0cm 0.0cm 0.0cm},clip,angle =0]{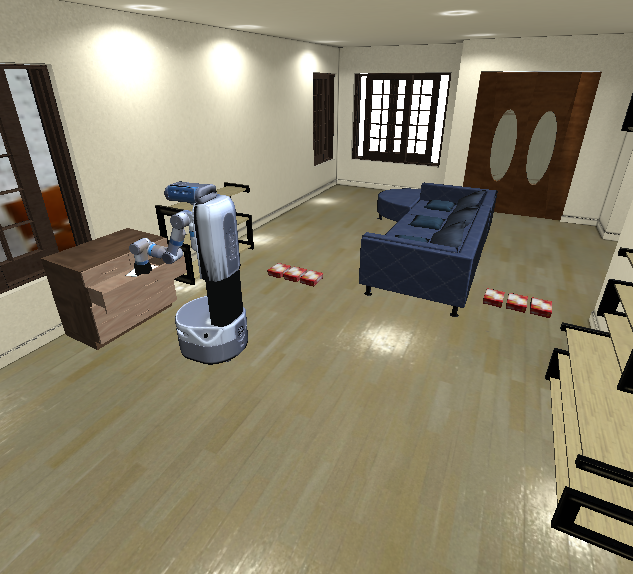}
	}
	\caption{In order to explore unstructured environments, the agent has to autonomously manipulate the environment which may include opening doors or looking into articulated objects.}
  	\label{fig:3rdperson}
\end{figure}

\figref{fig:3rdperson} depicts an example simulation environment.


\para{Doors:} Doors leading out of the apartments are set to a locked state and cannot be opened by the agent as they lead to a fall into the abyss.

\para{Cabinet positioning:} In the simulation, we first mark feasible areas along the walls of the apartment in which the cabinets can be spawned without blocking doorways or narrow corridors. We then uniformly sample poses from these areas and reject any poses that would result in a collision with the environment.

\para{Target object sampling:} During training, we draw six objects randomly with replacements. If the same object gets drawn repeatedly, the resulting object slot is left empty, leading to a distribution over 1-6 target objects. During the evaluation, the desired number of target objects is drawn uniformly from all target objects. Three out of the six target objects are set to always be placed in a drawer if they get selected. Leading to an average of half the target objects always being placed within a drawer.


\section{Interaction and Manipulation Motions}
\label{sec:motion_details}

As discussed in the assumptions in Section~IV, we assume the capability to infer the poses of objects of interest. For doors that can open to either side, this includes knowledge of the direction in which they open. All manipulation motions start with the agent navigating to a given offset point in front of the respective object. The opening motions in the simulation are then implemented in two variations: "magic" actions for fast training and physical execution based on a motion planner during evaluation.

\para{Magic actions:} During training, the object joint positions are slowly increased to their maximum collision-free value by the simulator. These magic opening actions can fail with a probability of 15\%, in which case the joint positions are left unchanged.

\para{Motion planning:} For realistic evaluation, at test time in simulation these motions are replaced with actual execution. The manipulation motions are implemented based on a BiRRT motion planner~\citeS{qureshi2015intelligent}. The motion planner creates a plan for opening the doors toward a desired direction. It selects a desired end-effector goal given the position and orientation of the object and plans a trajectory towards this goal by sampling collision-free arm joint configurations. Subsequently, the joints are set according to the plan. In addition, the end-effector closes to grasp the door knob. Finally, a push or pull operation manipulates the door by using the desired direction and computing the joint positions with inverse kinematics.

\begin{figure*}
	\centering
  		\includegraphics[width=1.0\linewidth,trim={0.0cm 0.0cm 0.0cm 0cm},clip,angle =0]{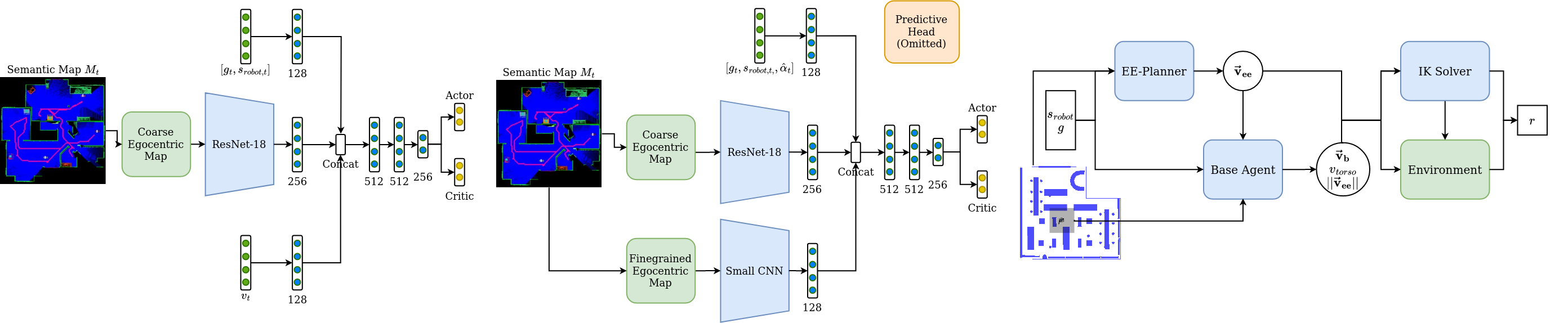}
	\caption{Network architectures of the high-level agent (left), the local exploration policy (middle) and the mobile manipulation policy (right). The Resnet-18 are pretrained on Imagenet. The predictive head of the low-level policy is omitted for clarity. $\hat{\alpha}_t$ is a vector of the angular current and previous predictions, $\vec{v}_{ee}$, $\vec{v}_{b}$ and $\vec{v}_{torso}$ are end-effector, base and torso velocities  $r$ is a reward signal. Full details of the local and mobile manipulation policies can be found in \cite{schmalst22exploration} and \cite{honerkamp2022learning}.}
  	\label{fig:networks}
\end{figure*}

For interacting with the cabinets and drawers in the real-world experiments we build upon a learning-from-demonstration framework \cite{twelsche2017learning}. In this framework, the robot observes trajectories of teacher hand demonstrations to accomplish the desired task. It then learns an encoding of a dynamical system representation of the trajectory via a Gaussian mixture model. We directly use these end-effector motions and transform then into mobile manipulation motions for the \textit{Toyota HSR} with the neural navigation for mobile manipulation approach ($N^2M^2$)~\cite{honerkamp2022learning}. $N^2M^2$ consists of a trained agent that generates velocities for a mobile base given a motion for the end-effector. This agent is trained on a random goal-reaching task and maintains kinematic feasibility between the end-effector and base while avoiding obstacles. The joint poses of the robot arm are then generated with an inverse kinematics solver based on the desired end-effector pose and the current base position. An overview of this approach is shown in \figref{fig:networks} (right).
However, the details of these manipulation policies are independent of our high-level reasoning approach, and in principle other imitation learning methods, such as ~\citeS{ijspeert2013dynamical, koskinopoulou2016learning} could be used to learn and provide the required manipulation behaviors.

\begin{table*}
    \centering
    \caption{Evaluation of exploration with and without geometric utility, reporting the success rate (top) and SPL (bottom). The models without geometric utility are the same as in Tab. I of the main text and reported here for direct comparison.}
    \label{tab:geometric}
    \begin{tabularx}{\textwidth}{cl|YYYYYY|Y||YYYYYY|Y}
      \toprule
        & Model & \multicolumn{7}{c||}{Seen} & \multicolumn{7}{c}{Unseen}\\
        \cmidrule{3-16}
        & & 1-obj & 2-obj & 3-obj &4-obj & 5-obj & 6-obj & Avg 1-6 & 1-obj & 2-obj & 3-obj &4-obj & 5-obj & 6-obj & Avg 1-6 \\
      \midrule
        \parbox[t]{3mm}{\multirow{6}{*}{\rotatebox[origin=c]{90}{Success}}} 
        & Greedy &  \textbf{96.8} & 94.0 & 91.0 & 91.3 & 89.7 & 88.0 & 91.8 & 96.4 & 94.9	& 92.6	& 93.7	& 91.4	& 87.2	& 92.7 \\
        & Greedy geometric &  95.5	&95.5	&93.0	&88.3	&88.8	&85.5 & 91.1 &97.0 & 95.4 & 94.3 & 95.2 & 91.4 & 92.6 & 94.3\\

        & SGoLAM+ &  96.0 & 93.7 & 90.2 & 85.8 & 88.0 & 85.3 & 89.8 & 95.6	&94.1	&94.7	&90.5	&91.6	&91.6	&93.0\\
        & SGoLAM+ geometric&  93.2 & 89.7 & 89.2 & 84.5 & 84.8 & 83.8 & 87.5 & 95.0 & 92.2 & 93.0 & 89.9 & 89.9 & 90.9 & 91.8\\
        & \ours{} &  \textbf{96.8}	&\textbf{96.5}	&94.0	&\textbf{94.8}	&\textbf{93.5}	&\textbf{91.7}	&\textbf{94.6} & \textbf{97.5}	&\textbf{97.9}	&\textbf{95.6}	&\textbf{95.6}	&\textbf{93.5}	&\textbf{93.5}	&\textbf{95.6} \\
        & \ours{} geometric &  96.8	&93.5	&\textbf{96.5}	&91.8	&91.0	&90.3	&93.3 & 97.9	&97.1	&96.2	&94.3	&95.0	&90.5	&95.2 \\
        
        \midrule
        \parbox[t]{3mm}{\multirow{6}{*}{\rotatebox[origin=c]{90}{SPL}}}
        & Greedy &  43.1 & 45.3 & 48.6 & 50.8 & 54.1 & 56.9 & 49.8 &45.0	&49.5	&51.8	&57.9	&60.6	&62.0	&54.5\\
        & Greedy geometric&  41.9 & 44.7 & 48.2 & 49.2 & 52.3 & 53.2 & 48.2 & 44.4 & 46.4 & 52.3 & 55.0 & 58.9 & 61.0 & 53.0\\
        
        & SGoLAM+ &  41.5 & 44.3 & 47.8 & 48.9 & 52.5 & 55.3 & 48.4 & 44.8	&49.1	&54.2	&56.2	&60.3	&64.2	&54.8\\
        & SGoLAM+ geometric& 42.5 & 41.7 & 44.3 & 46.8 & 48.8 & 52.1 & 46.0 & 44.0 & 46.6 & 50.6 & 52.7 & 58.7 & 62.1 & 52.5\\
        & \ours{} &  \textbf{49.7}	&\textbf{51.5}	&\textbf{55.2}	&\textbf{57.5}	&\textbf{59.5}	&\textbf{60.9}	&\textbf{55.7} & \textbf{52.2}	&\textbf{56.0}	&\textbf{57.6}	&\textbf{63.5}	&\textbf{65.7}	&\textbf{68.7}	&\textbf{60.6}\\
        & \ours{} geometric & 47.8	&47.9	&54.7	&54.0	&54.8	&58.6	&53.0 & 49.2	&51.3	&55.9	&58.8	&64.2	&65.1	&57.4 \\
      \bottomrule

    \end{tabularx}
\end{table*}

\setlength{\tabcolsep}{6pt}
\begin{table}
  \centering
  \caption{Hyperparameters used for training.}
  \label{tab:hyper}
  
  \begin{tabular}{l|c|l|c}
  \toprule
Parameter & Value & Parameter & Value \\
\midrule
clip param      & 0.1      & $\gamma$        & adaptive\\
ppo epoch       & 4        & learning rate              & 0.0005\\
num mini batch  & 128      & optimizer       & Adam \\
entropy coef    & 0.005    & &\\

\bottomrule

  \end{tabular}
  
\end{table}

\section{Navigation Motions}

As discussed in Section~IV, the navigation policy moves the agent along waypoints computed by an A* planning algorithm. The algorithm computes the path based on a prior known traversability map with an inflation radius of \SI{0.2}{\meter}. This map is used simply to avoid recomputing the navigation graph at every step. 
For frontier point selection, the semantic map is first converted to an occupancy map and then convolved with a $5\times5$ kernel for a single iteration.

\section{Real-World Adaptations}
\label{sec:real_world_details}

In contrast to the Fetch robot that we trained in simulation, the HSR has an omnidirectional drive. The pre-trained local exploration policy still executes its commands as pure differential drive motions (sending forward and angular velocity commands). The unseen $N^2M^2$ mobile manipulation policy and the ROS navigation module can make use of the robot's omnidirectional movement, as the training procedure is agnostic to their internal workings. To account for differences in the robot geometry, we use a robot-specific inflation radius for the navigation policies and adjust the instance navigation module to select relative navigation goals that are further away from the object instances.

\section{Deployment with a Full Perception Pipeline}
\label{sec:real_world_deployment}

In real-world experiments, we provide the agent with a pre-annotated semantic map to match the assumption of accurate semantic perception. While we consider evaluating the entire system with a full perception pipeline as future work, this section provides details on the requirements and a possible implementation to facilitate the deployment.

Real-world deployment requires (i) an RGB-D sensor (ii) visual localization and mapping (iii) semantic segmentation and (iv) grasp pose detection. (i) and (ii) can be achieved with modern RGB-D SLAM approaches such as RTAB-Map~\citeS{labbe2019rtab}. For semantic segmentation, a wide range of models exists, including older methods such as Mask R-CNN~\citeS{he2017mask} as well as newer, transfomer-based methods~\citeS{cheng2021mask2former, kappeler2023few}. The best model is use-case dependent and should be chosen based on the available compute, the required object categories, and the required accuracy. We demonstrate in our real-world experiments that a model that can accurately detect target objects, doors, and articulated objects is sufficient, as we map all other categories simply to the wall category and find that the model is robust to this remapping. We hypothesize that further improvements can be achieved by training the agent with the imperfect outputs from the semantic model to condition it on the strengths and weaknesses of a particular model, which we aim to evaluate in future work

For a possible implementation, users can follow the method of~\citeS{chen2023not}: deploy an RTAB-node for localization and mapping. To extend it to semantic labels, deploy a second RTAB-node that listens to the semantic masks. Then fuse the resulting point clouds.
Finally, for handle detection and grasp-pose detection, \citeS{arduengo2021robust} achieve accurate results with a retrained YOLO model on a public handle-specific dataset.

\section{Training Details}
\label{sec:training_details}  

\para{Network architectures:} 
The network architectures of the high-level policy, the local exploration policy~\citeS{schmalst22exploration} and the $N^2M^2$-policy~\citeS{honerkamp2022learning} are shown in \figref{fig:networks}.
The coarse map is encoded into a 256-dimensional feature vector with a ResNet-18~\citeS{he2016deep}. The fine-grained map is encoded into a 128-dimensional feature vector using a simple three-layer CNN with 32, 64, and 64 channels and strides 4, 2, and 1. The local exploration policy then concatenates the map encodings together with the robot state and processes these features with fully connected layers, following the author's original architecture~\citeS{schmalst22exploration}.
The high-level policy only uses the coarse map encoder architecture (without weight-sharing). Both the local exploration policy and the high-level policy then use an actor and a critic parameterized by a two-layer MLP network with 64 hidden units each. The high-level agent learns a discrete, Categorical policy while the local exploration policy is parameterized as a Gaussian policy.
The mobile manipulation policy receives a local occupancy map created from its LiDAR scanner. It encodes the map at two resolutions, then concatenates these features with the robot state and passes it through three fully connected layers to produce actions for the base and torso of the robot as well as the norm of the end-effector velocities.

\para{Hyperparameters:} \tabref{tab:hyper} lists the main hyperparameters used during training. The agents were implemented based on a public library~\citeS{stable-baselines3}. Parameters not mentioned were left at their defaults. The $\gamma_{adaptive}$ parameter is calibrated to result in an average discount factor of $0.99$ for a high-level policy step. This is done by setting $\gamma_{adaptive}^n = 0.99$ where $n$ is the average subpolicy duration. We set $n$ to $7$ for  \ours{} and \textit{w/o frontier} and 10 for \textit{w/o expl} based on average subpolicy lengths in a training run. During training, we early terminate the episodes if they exceed 500 steps.

\section{Ablation: Geometric Utility Baseline}

In this section, we compare with recent work on semantic object-navigation~\cite{zheng2022towards, zhou2023esc, chen2023not}. As our task definition builds on previous work that focuses on random target object placements~\cite{wani2020multion, schmalst22exploration}, there are no correlations between objects that could be exploited. So to compare with these methods we focus on the non-semantic components and implement the geometric utility for frontier selection proposed by recent work~\cite{chen2023not} within the global exploration subpolicy. This enables us to extend this work, which focuses on non-interactive search, to our proposed interactive search.

We evaluate all the models with this new frontier selection. The results are reported in \tabref{tab:geometric}, together with the results for the original versions of each model for direct comparison.
While the geometric utility leads to small improvements on a subset of the tasks, most notably it increases the average success rate for the greedy model on unseen scenes, overall it leads to a small decrease in both the success rate and the SPL across all other settings.

{\footnotesize
\bibliographystyleS{IEEEtran}
\bibliographyS{biblio}
}






\end{document}